\documentclass[runningheads]{llncs}

\usepackage{accv}

\usepackage{accvabbrv}
\usepackage{graphicx}
\usepackage{booktabs}
\usepackage{amsmath}
\usepackage{amssymb}
\usepackage{etoolbox}
\usepackage[accsupp]{axessibility}

\usepackage[pagebackref,breaklinks,colorlinks,citecolor=accvblue]{hyperref}

\usepackage{color, colortbl}
\usepackage{multirow}
\usepackage{threeparttable}
\usepackage{xspace}
\usepackage{makecell}
\usepackage{pifont}

\newcommand{\R}[1]{{%
    \textbf{%
        \ifstrequal{#1}{1}{\textcolor{red}{R#1}}{%
        \ifstrequal{#1}{2}{\textcolor{blue}{R#1}}{%
        \ifstrequal{#1}{3}{\textcolor{magenta}{R#1}}{%
        \ifstrequal{#1}{4}{\textcolor{teal}{R#1}}{%
                           \textcolor{cyan}{R#1}%
        }}}}%
    }%
}}

\colorlet{colorFst}{Green!25}       
\colorlet{colorSnd}{Green!10} 
\colorlet{colorTrd}{Yellow!15}      
\colorlet{colorLow}{darkgray!30}    
\colorlet{nogood}{Red!30}    
\definecolor{ours}{rgb}{0.9, 0.95, 0.94}
\definecolor{darkgreen}{rgb}{0.00, 0.8, 0.2}
\definecolor{darkyellow}{rgb}{0.96, 0.75, 0.00}
\definecolor{badcolor}{rgb}{0.82,0.25,0.12}

\newcommand{\ours}{\cellcolor{ours}}
\newcommand{\ds}{\cellcolor{colorSnd}}

\newcommand\boldblack[1]{\textcolor{black}{\textbf{#1}}}

\def\eg{\emph{e.g.}\xspace} 
\def\ie{\emph{i.e.}\xspace}

\begin{document}

\title{NaCR: Visual Localization via NeRF-aided Camera Ray Regression}
\titlerunning{NeRF-aided Camera Ray Regression}

\author{Yesheng Zhang, Xiang Dai, Xu Zhao, Chongyang Zhang}
\authorrunning{Y.~Zhang}
\institute{Shanghai Jiao Tong University}

\maketitle

\begin{abstract}
Visual localization (VL) is a fundamental technology for vision applications such as virtual reality. Recently, a novel VL paradigm, Camera Ray Regression (CRR), has emerged, which maps 2D image patches to 3D camera rays, but its accuracy is limited. To improve CRR accuracy, we notice a compelling duality: the inverse of this mapping is inherently performed by the novel view synthesis model, \ie, Neural Radiance Fields (NeRF). While NeRF renders image patches from camera rays via differentiable ray marching, CRR predicts the rays from image patches. Motivated by this complementary relationship, we propose NeRF-aided Camera Ray Regression (NaCR), a unified framework that seamlessly bridges NeRF and CRR at the ray level. First, NaCR incorporates three simple yet effective enhancements into the CRR baseline. Second, leveraging a pre-trained NeRF, NaCR augments the training data by synthesizing novel views tailored for efficient, patch-level consumption. Finally, exploiting the differentiability of NeRF, NaCR forms a closed-loop supervision pipeline where photometric rendering errors are back-propagated to optimize the predicted camera rays. To ensure stable convergence within the highly non-convex image space, we introduce a two-stage training curriculum. Extensive experiments across indoor and outdoor benchmarks demonstrate that NaCR achieves competitive accuracy. Comprehensive ablation studies validate the efficacy of each proposed component.
\end{abstract}

\section{Introduction}
\label{sec:intro}

Visual Localization (VL) targets estimating the 6-Degree-of-Freedom camera pose for a given query image in a known scene. It has a variety of applications, such as SLAM/SfM~\cite{orb}, Virtual Reality (VR), and Augmented Reality (AR)~\cite{acezero}. Although it is a long-standing research area, achieving high-precision visual localization remains a consistent pursuit.

\begin{figure}[!t] 
    \centering
    \includegraphics[width=0.5\linewidth]{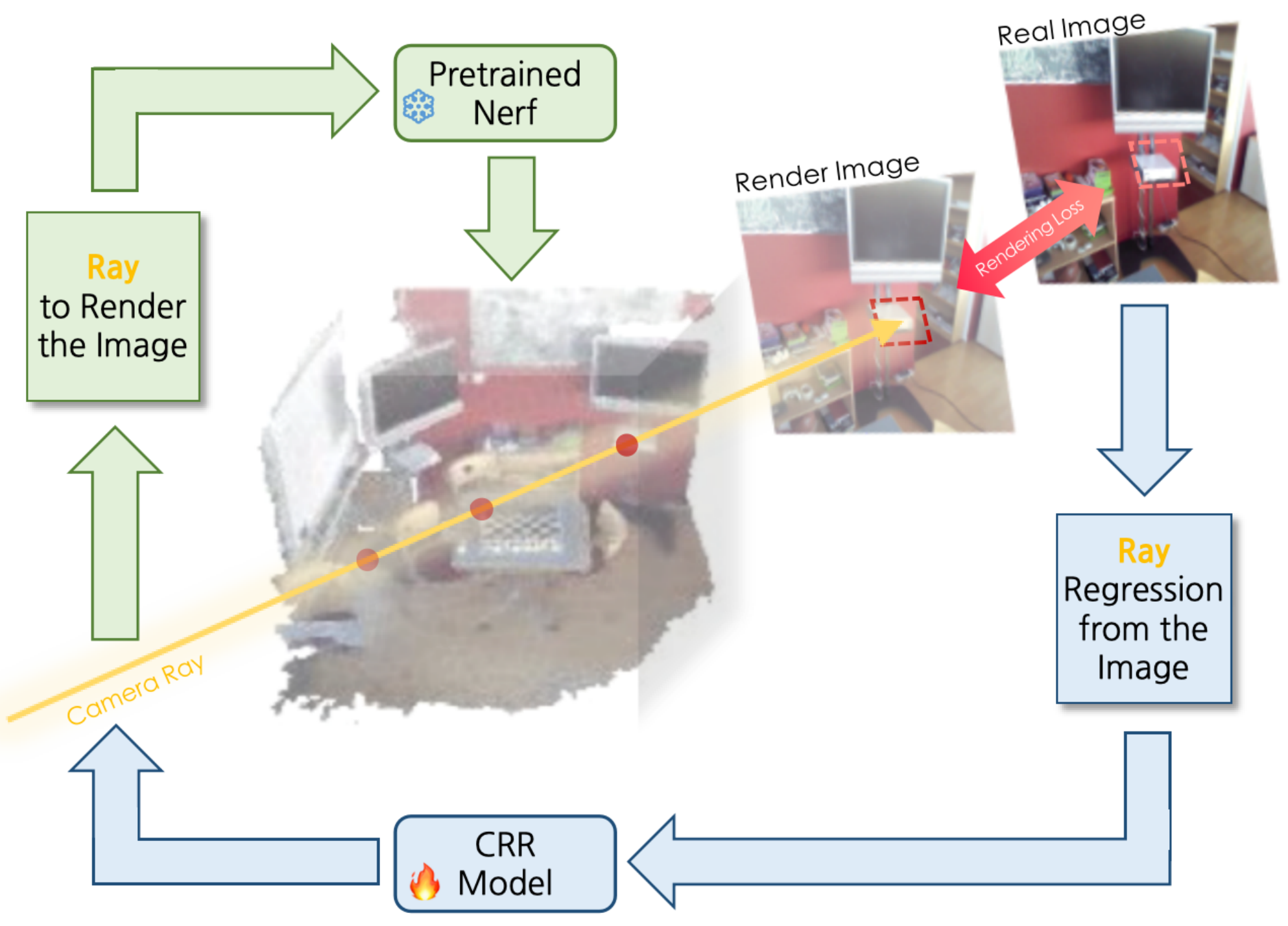} 
    \caption{Camera Ray Regression (CRR) and Neural Radiance Fields (NeRF) Represent Two Complementary Processes: { \normalfont CRR learns to map an image patch to a \colorbox{colorTrd}{camera ray}, while NeRF renders an image patch from a \colorbox{colorTrd}{camera ray}. Recognizing this synergy, we introduce NeRF into the CRR training pipeline, creating an end-to-end differentiable loop. In this loop, photometric rendering error provides a direct guidance for CRR optimization. Additionally, NeRF allows for generating synthetic training image patches from arbitrary rays, serving as a powerful patch-level data augmentation technique.}}
    \label{fig:first}
\end{figure}

In the era of deep learning, end-to-end VL methods employ neural networks as implicit maps to predict camera poses for query images. 
They are computationally more efficient than conventional VL methods~\cite{b2f,hloc}, which require storing explicit 3D maps. 
As a representative end-to-end paradigm, Scene Coordinate Regression (\textbf{SCR})~\cite{dsac} introduces an \textit{over-parameterization} trick for VL. It first estimates dense scene coordinates corresponding to image patches, and then calculates the pose based on RANSAC-PnP methods. Among the SCR series,  ACE \cite{ace} achieves efficient model optimization through Gradient Decorrelation Training~(\textbf{GDT}), yielding precise results. Nevertheless, the spatial distribution of 3D scene points exhibits high variance (especially background points), leading to unstable training. Thus, SCR methods~\cite{ace-i2,ace-i0,GLACE} heavily rely on depth priors and hyperparameter tuning to mask out unstable predictions during training.

To eliminate this hyperparameter restriction, DIMM~\cite{Zhang_2025_ICCV} proposes a novel VL paradigm, \ie, Camera Ray Regression (\textbf{CRR}). In this paradigm, the model regresses a camera ray for each image patch. 
As a patch-level pose representation, camera rays are naturally over-parameterized and compatible with GDT.
Bounded ray errors yield robust training dynamics, thus eliminating the need for depth priors.
After ray regression, the camera pose can be recovered by two linear solvers for rotation and translation, respectively. This decoupled approach is more robust than the PnP solvers in SCR. Collectively, these benefits of \textit{camera rays} enable CRR-based methods~\cite{Zhang_2025_ICCV,grloc} to achieve promising VL accuracy in outdoor scenes. However, their accuracy remains limited, particularly in indoor scenes, possibly due to constrained training scalability and reliance on purely geometric supervision.

Early on, camera rays were utilized as the medium for rendering in the novel view synthesis (NVS) model: NeRF~\cite{nerf}. Its Ray Marching process establishes a differentiable pathway from rays (poses) to NeRF and further to images, which can be integrated in VL \textit{inference} for pose optimization~\cite{barf,nerfmatch,posefromnerf}. However, recalling CRR, we recognize that it performs the exact inverse operation (see \cref{fig:first}): a learned, differentiable mapping from an image patch back to its corresponding camera ray. 
By integrating NeRF into the CRR \textit{training}, the gradient propagation can realize a closed loop from images to rays and back to images. Then, the photometric error from NeRF's rendering provides a powerful, self-supervised signal that directly optimizes the CRR network predictions.
\textit{In essence, this provides a mechanism to align the implicit scene representation learned by the CRR model with the volumetric scene representation encoded in NeRF.}
This complementary relationship is beneficial for improving CRR accuracy. 
Furthermore, this integration is highly practical. Because NeRF renders on a ray-level basis, it seamlessly integrates with the patch-level, ray-based GDT~\cite{ace} of CRR~\cite{Zhang_2025_ICCV}.

\begin{figure*}[!t] 
    \centering
    \includegraphics[width=0.9\linewidth]{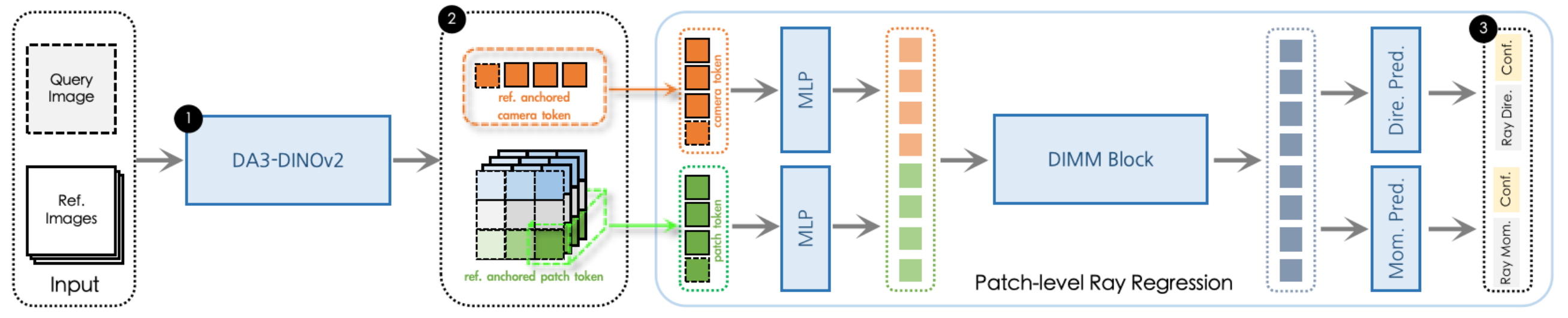} 
    \caption{The Improved DIMM. \normalfont Our work introduces three \textit{simple yet effective} enhancements to DIMM. \ding{182} We substitute the original DINOv2 trained on 2D tasks with the DA3-DINOv2 backbone pre-trained for 3D reconstruction, which provides the model with richer geometric priors. \ding{183} We transform the challenging absolute camera ray regression into a more tractable task by conditioning the model on tokens from fixed reference views. \ding{184} We augment the DIMM output head to predict confidence for each ray. This effectively leverages the over-parameterized nature of camera rays by down-weighting uncertain predictions.}
    \label{fig:i-dimm}
\end{figure*}

The integration of NeRF also provides a native data augmentation mechanism for CRR. Since end-to-end VL requires training data of \textit{image-pose} pairs, using NVS models to expand this dataset is a direct choice. Indeed, recent work has demonstrated that augmenting VL training with randomized-appearance views generated by 3D Gaussian Splatting (\textbf{3DGS})~\cite{3dgs} can significantly boost accuracy~\cite{rap,grloc}. As mentioned above, a pre-trained NeRF can be a supervised module in CRR training, using it for data augmentation incurs minimal overhead. Furthermore, CRR can leverage this synthetic data with superior efficiency. Unlike current pose-level training\cite{rap,grloc}, where an entire image supports a single pose prediction~\cite{rap}, CRR training consumes these synthetic views at the ray (patch) level. This enables a more efficient use of the augmented data, maximizing the benefit of generated views.

Thus, we propose \textbf{N}eRF-\textbf{a}ided \textbf{C}amera \textbf{R}ay Regression, namely \textbf{NaCR}, a novel framework that tightly integrates NeRF and CRR at the ray level to achieve accurate VL. We first introduce three \textit{simple yet effective} architectural improvements to the baseline CRR model (DIMM), inspired by recent advances in ray-based 3D reconstruction~\cite{depthanything3}. Subsequently, NaCR incorporates NeRF into the CRR training, serving two synergistic purposes. \textbf{1) Data Augmentation}: NaCR exploits NeRF to render novel views, which are comprehensively consumed at the patch level during CRR training. To mitigate the domain gap between real and synthetic images, we also incorporate adversarial loss in training~\cite{rap}. \textbf{2) Closed-Loop Supervision}: NaCR leverages the differentiability of NeRF to enforce closed-loop consistency. This enables the patch-level photometric rendering loss to supervise ray-level regression, guiding the CRR model to align with the robust 3D scene priors embedded in NeRF. However, naively optimizing rendering loss from scratch can collapse the CRR model due to the highly non-convex nature of the image space~\cite{posefromnerf}. Thus, we propose a two-stage training pipeline. In the first stage, the CRR model is optimized using standard ray-confidence loss alongside NeRF-based data augmentation. Once the CRR performance stabilizes, the second stage introduces the rendering loss. This strategy allows NeRF's image-space supervision to gracefully fine-tune the regressed rays within a reliable local neighborhood, yielding significant accuracy gains. 

In summary, our contributions are as follows:
\begin{enumerate}
    \item  We present NaCR, a novel visual localization framework that tightly integrates Neural Radiance Fields (NeRF) and Camera Ray Regression (CRR) at the ray level. NaCR harnesses NeRF as both a dense patch-level data augmenter and a differentiable module providing closed-loop photometric supervision for ray regression.
    \item We introduce three simple yet highly effective architectural enhancements over the baseline DIMM network, improving the ray regression precision.
    \item Extensive experiments on indoor and outdoor benchmarks demonstrate that NaCR improves the CRR accuracy and achieves comparable performance in VL. Comprehensive ablation studies are provided to validate the efficacy of each proposed component.
\end{enumerate}

\section{Related Work}
\label{sec:rw}

\subsection{Visual Localization}
Early localization methods\cite{b2f,hloc} matched query features~\cite{match0,match2} to explicit 3D point-cloud maps and recovered poses with PnP. 
The storage cost of such maps has motivated learning-based methods~\cite{pn0,ace,ace-i2,GLACE} that encode scenes in network parameters.
These methods mainly follow three paradigms. Absolute and Relative Pose Regression (APR/RPR) directly predict compact poses from query images; they are efficient but noise-sensitive and often less precise. Scene Coordinate Regression (SCR) predicts an over-parameterized pose representation, \ie, scene coordinates for image patches. Although accurate, SCR uses unbounded coordinates, which can destabilize training and require depth priors for outlier filtering. Camera Ray Regression (CRR) instead predicts camera rays, whose bounded errors enable smoother training without carefully tuned hyperparameters. We build on CRR and use camera rays as a geometric bridge to incorporate Neural Radiance Fields (NeRF), providing both synthetic data augmentation and robust image-level supervision.

\subsection{Camera Ray Representation}
Camera ray representation lifts a compact camera pose matrix to an image-aligned ray map. This dimensional consistency makes rays useful across pose-centric 3D vision tasks. In novel view synthesis (NVS), rays serve as view conditions~\cite{lvsm25,rayzer,erayzer} or rendering media~\cite{nerf,nerfstudio}; in dense prediction, ray maps are often used as targets~\cite{ray-calib,ray-depth,raydiffusion}. 
Recent large-scale 3D reconstruction models~\cite{depthanything3,matrix3d,diffusionsfm} show that data-driven models can learn generalizable mappings from image patches to camera rays. Camera rays also exhibit favorable convergence in optimization-based 3D reconstruction~\cite{glomap,mast3r-slam}. For VL, rays share patch-level separability with scene coordinates~\cite{grloc}, enabling scene-wise Gradient Decorrelation Training and competitive localization performance~\cite{Zhang_2025_ICCV}. We leverage this geometric and dimensional alignment to closely couple CRR with a NVS rendering pipeline (NeRF).

\subsection{Novel View Synthesis in Visual Localization}

Novel View Synthesis (NVS) models such as NeRF~\cite{nerf} and 3DGS~\cite{3dgs} can render high-fidelity images from arbitrary viewpoints, providing dense image-pose pairs for VL training. Their differentiability also allows photometric errors from rendered images to update pose parameters.
NVS has therefore been used in VL in two mostly decoupled roles. \textbf{1)} Matching-based methods integrate NVS into pose refinement~\cite{posefromnerf}: CrossFire~\cite{crossfire} uses NeRF for iterative rendering and pose optimization, and NeRFMatch~\cite{nerfmatch} propagates photometric errors to a pose optimization module. \textbf{2)} End-to-end methods use NVS for scene-specific data augmentation: RAP~\cite{rap} and GRLoc~\cite{grloc} render appearance-varied novel views with 3DGS to enrich training data and improve localization.
Our method also introduces NeRF into VL, but differs by using camera rays to tightly couple NeRF with CRR. Rather than treating NeRF only as an offline augmentation engine or post-hoc refinement module, we exploit both its data generation and ray-level differentiability during CRR training. 

\section{Methodology}
\label{sec:m}

\begin{figure*}[!t]
    \centering
    \begin{subfigure}[t]{0.49\linewidth}
        \centering
        \includegraphics[width=\linewidth]{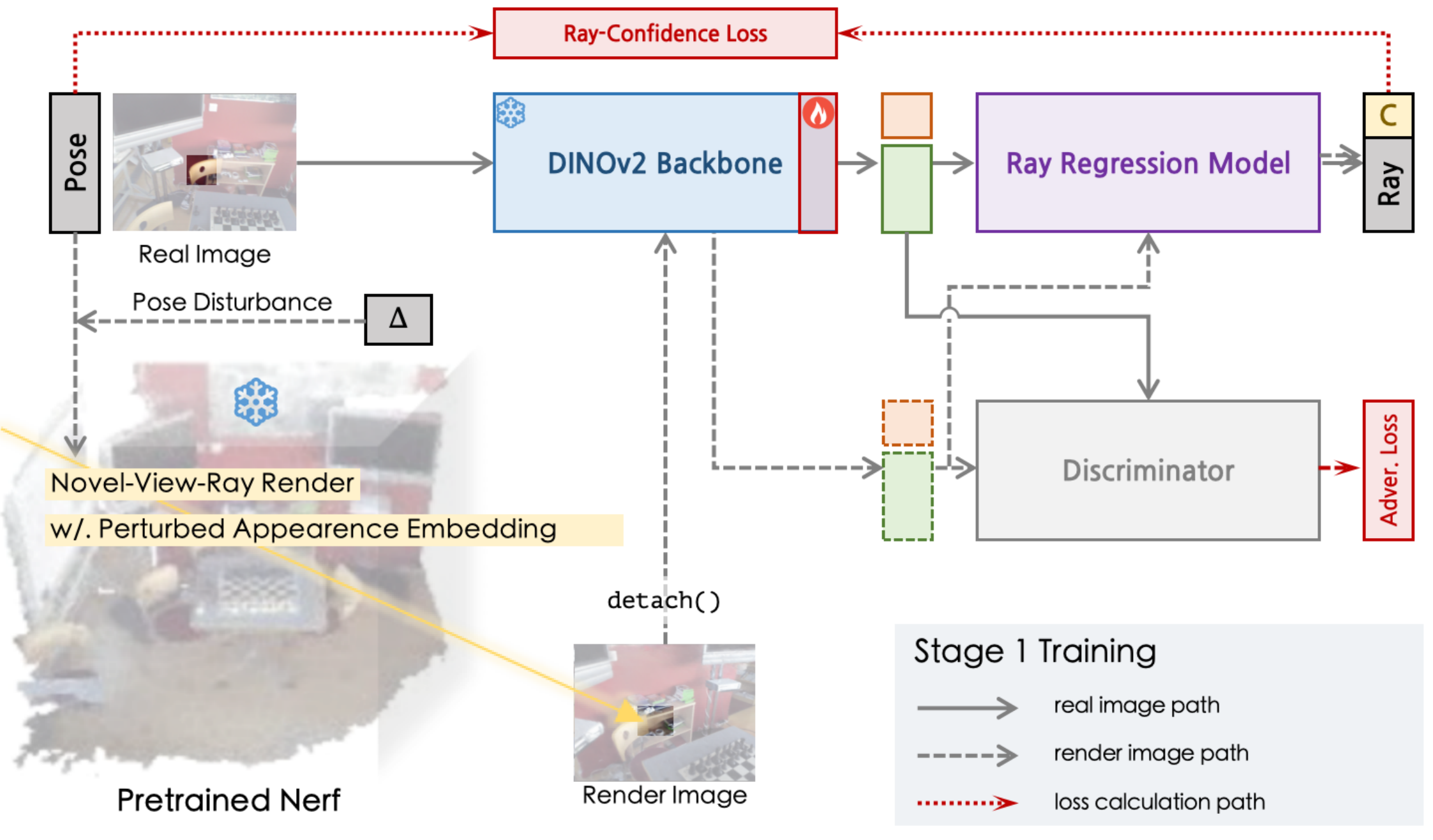}
        \caption{Training Stage 1.}
        \label{fig:ts1}
    \end{subfigure}
    \hfill
    \begin{subfigure}[t]{0.48\linewidth}
        \centering
        \includegraphics[width=\linewidth]{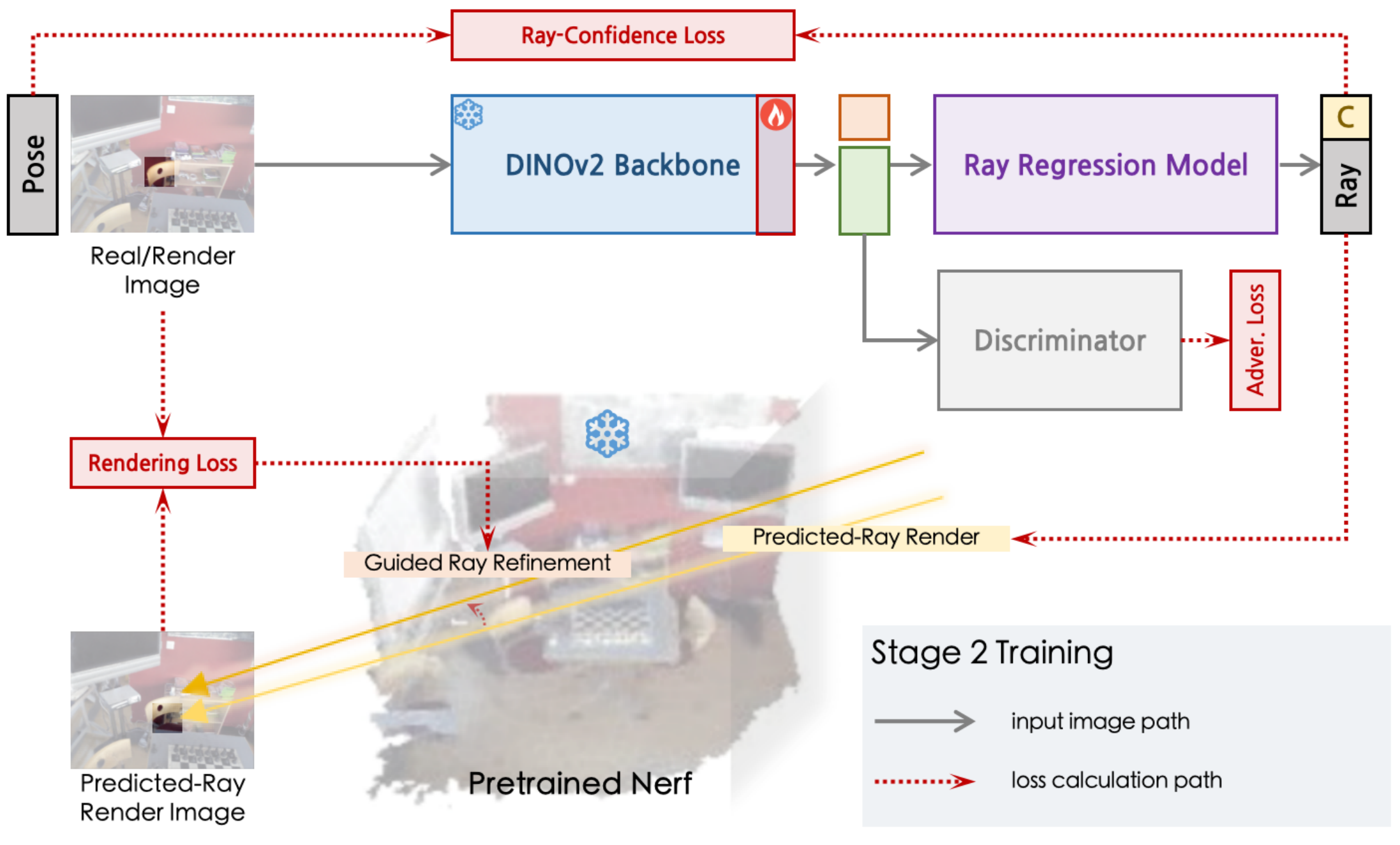}
        \caption{Training Stage 2.}
        \label{fig:ts2}
    \end{subfigure}
    \caption{The two-stage NeRF-aided CRR training pipeline. \normalfont (a) Stage 1 trains CRR with patch-level Gradient Decorrelation Training (GDT) on real and NeRF-synthesized patches, while adversarial feature alignment bridges the render-to-real gap. (b) Stage 2 adds a differentiable rendering loss, using the photometric error between rendered and original patches as a closed-loop signal for ray refinement.}
    \label{fig:training-stages}

\end{figure*}

\subsection{Preliminaries}

\noindent\textbf{{Camera Ray Regression (CRR).}}
A camera ray is defined as the ray originating from the camera center and passing through an image pixel~\cite{raydiffusion}. A minimum of two camera rays can determine the camera origin, while the direction of a camera ray can establish the camera rotation. Therefore, the set of camera rays determined by the centers of uniformly divided image patches (\ie, the ray map) can serve as an over-parameterized representation of the camera pose. Recently, this representation has proven pivotal in 3D vision tasks such as Novel View Synthesis~\cite{lvsm25,rayzer,erayzer}, 3D Reconstruction~\cite{depthanything3,diffusionsfm,matrix3d}, and Visual Localization~\cite{Zhang_2025_ICCV,grloc}.
DIMM~\cite{Zhang_2025_ICCV} pioneers the CRR paradigm by designing an end-to-end framework to estimate patch-level camera rays for query images in visual localization.
Specifically, for an input image $I \in \mathbb{R}^{W \times H \times 3}$, it first extracts patch-level image feature tokens $\mathcal{T}_{i} \in \mathbb{R}^{D}$, where $D$ is the feature dimension, corresponding to image patches $p_i \in \mathbb{R}^{s \times s \times 3}$ of size $s$, with $i = \{1, \dots, N\}$ and $N = WH/s^2$. Then, the model regresses the patch-level camera rays $r_i \in \mathbb{R}^6$. Each ray has six parameters. The first three are ray direction $rd_i \in \mathbb{R}^3$ and $||rd_i||_2=1$. The last three are ray moment $rm_i \in \mathbb{R}^3$.
After obtaining the predicted rays $\hat{\mathcal{R}} = \{r_i\}_{i=1}^N$, rotation and translation can be solved separately through two linear problems~\cite{raydiffusion}, thereby achieving VL for the query image.

\noindent\textbf{{Baseline DIMM.}}
The DIMM model primarily consists of a feature extractor $\mathcal{E}$ and a ray regression model $\mathcal{F}$. It first employs a fine-tuned DINOv2 as the feature extractor to acquire patch-level feature tokens $\{\mathcal{T}_i\}_{i=0}^N$, each formed by concatenating local features, global features, and image-location features~\cite{Zhang_2025_ICCV}:
\begin{equation}
\{\mathcal{T}_i\}_{i=0}^N = \mathcal{E}(I),
\end{equation}
Then, for each feature token, the corresponding ray is achieved by:
\begin{equation}
r_i = \mathcal{F}(\mathcal{T}_i).
\end{equation}
$\mathcal{F}$ includes a parallel multi-mapper and a semantic attention block. Please refer to~\cite{Zhang_2025_ICCV} for more details.
During training, DIMM maps a single scene via Gradient Decorrelation Training (\textbf{GDT}) to model the spatial information of camera rays into $\mathcal{F}$.
In localization, it performs patch-level ray regression on the target image and then obtains the camera pose using two decoupled RANSAC-based solvers.
Our method builds upon this, making three simple improvements to the original DIMM architecture and introducing NeRF into the training pipeline.

\subsection{Improved DIMM}
Next, we elaborate our three improvements over the baseline DIMM (\cref{fig:i-dimm}): the utilization of a DA3-pretrained DINOv2 backbone, Reference Token conditioning, and a Ray-Confidence head.

\noindent\textbf{{Pretrained DINOv2 from DA3.}} Recently, the Depth Anything 3 (DA3)~\cite{depthanything3} framework demonstrated highly accurate and generalized camera ray regression with a concise DINOv2 backbone. Trained on massive data, DA3 achieves robust geometric priors of camera rays. Thus, transferring this knowledge for scene-wise mapping is beneficial for VL.
However, since the DA3's DPT decoder predicts rays at the image level, it is fundamentally incompatible with the spatially decoupled {GDT}~\cite{ace}, which is essential for fast convergence and high accuracy on individual scenes. Therefore, to harness the powerful priors from DA3 without sacrificing the efficiency of GDT, we adopt a hybrid approach: using the pre-trained DINOv2 backbone from DA3 while retaining DIMM's patch-level decoder.

\noindent\textbf{Reference Token Conditioning.} The CRR task involves estimating absolute ray parameters within a fixed scene. Thus, the model needs to learn the scene's reference coordinate system during training to fit the ground truth. However, our DINOv2 backbone from DA3, pre-trained for multi-view reconstruction, naturally operates in a relative frame of reference, conditioning its features on a reference view. Inspired by recent RPR methods~\cite{ff}, we fix the reference frame by inputting $F$ reference images to the DINO feature extractor. Thus, the query image features can be computed relative to the fixed reference images, and each target patch token $\mathcal{T}_i$ is concatenated with its corresponding reference tokens $\{\mathcal{T}^j_i\}_{j=0}^F$. This provides the CRR model with explicit reference information, thereby simplifying its learning task.
\begin{equation}
r_i = \mathcal{F}(\mathcal{T}_i ~|~ \{\mathcal{T}^j_i\}_{j=0}^F),
\end{equation}
where $i$ is the patch index and $j$ is the reference image index.
In practice, we empirically use $3$ reference images by clustering the training poses and choosing the $3$ cluster centroids.

\noindent\textbf{Ray Confidence Head.} DIMM directly predicts ray parameters without considering their confidence, forcing the model to treat all predictions as equally reliable. To address this limitation, and inspired by DA3~\cite{depthanything3}, we introduce a ray confidence head to DIMM. This mechanism effectively leverages the over-parameterized nature of camera rays by actively down-weighting the influence of uncertain predictions, leading to a more robust final pose estimate. 
At inference time, the predicted confidence scores are also used to weight the RANSAC-based ray solver~\cite{Zhang_2025_ICCV,depthanything3}, enhancing the robustness of the final pose.

\begin{figure*}[!t] 
    \centering
    \includegraphics[width=0.85\linewidth]{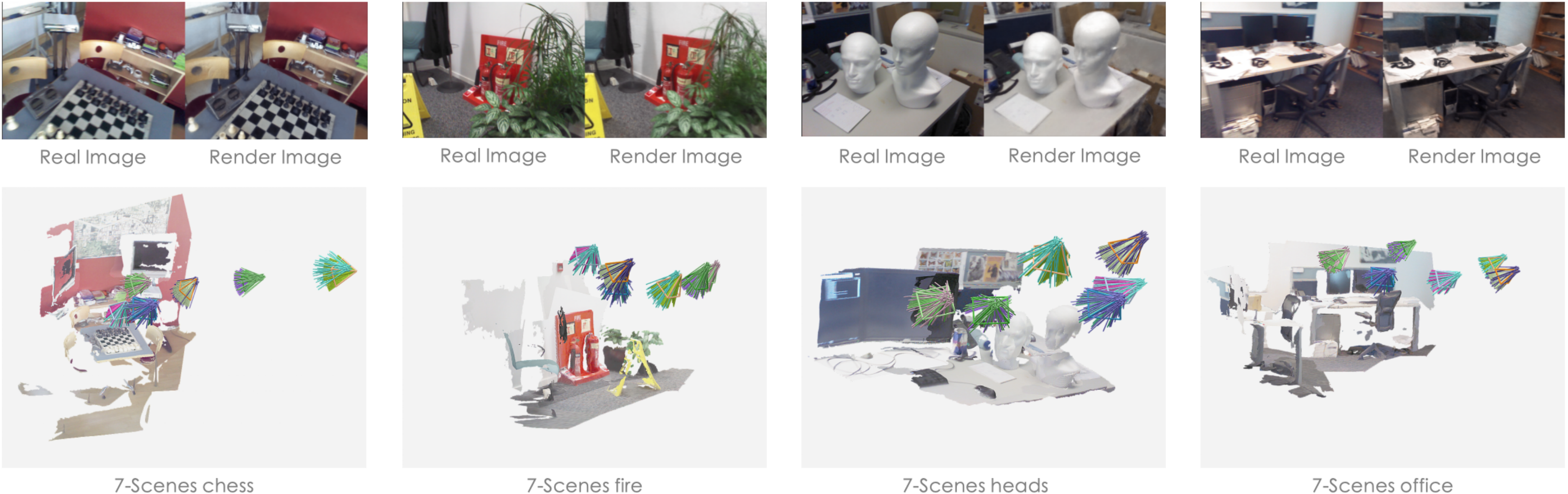} 
    \caption{Qualitative results in 7-Scenes. \normalfont The top row compares the ground truth image with a view rendered from our estimated camera pose, which validates the precision of our method. The bottom row visualizes the predicted and GT camera rays within the 3D scene.}
    \label{fig:q7s}
    \vspace{-0.2em}
\end{figure*}

\subsection{NeRF-aided CRR}

In this section, we introduce the NeRF integration of NaCR, which serves two primary purposes: providing dense data augmentation via novel view synthesis, and enabling self-supervised ray refinement through a differentiable closed-loop renderer. To effectively achieve this, we address two critical challenges, \ie, the render-to-real domain gap and non-convex photometric ambiguities, by introducing adversarial feature alignment and a robust patch-level rendering loss within a two-stage training paradigm.

\noindent\textbf{Motivation and Challenges.}
While recent works~\cite{rap,grloc} have shown that integrating NVS models for data augmentation in VL significantly improves accuracy, this benefit can be amplified in the context of CRR. As CRR treats each camera ray as an independent sample~\cite{Zhang_2025_ICCV}, a single rendered image provides a dense set of training data, maximizing data efficiency. Furthermore, the integration extends beyond augmentation. NeRF's ray-marching rendering is natively compatible with CRR's output, allowing us to connect the two models. This creates a differentiable, closed-loop pipeline where photometric rendering loss can be back-propagated to provide a powerful supervisory signal for the ray regressor. This signal guides the CRR model to become consistent with the implicit 3D scene representation encoded in NeRF.
However, a naive integration faces two major challenges: the render-to-real gap, and the non-convex photometric ambiguities.

\noindent\textbf{Bridging the Render-to-Real Gap.}
Firstly, the rendered images inherently contain noise compared to real images, which negatively impacts VL model training. To mitigate this, we follow RAP~\cite{rap}, using generative adversarial training to align the features generated from real and rendered images. Since we use the pre-trained DA3 DINO backbone, fine-tuning all its parameters is computationally expensive and risks destroying its semantic priors. Hence, we only unfreeze the last two attention layers of the DINO backbone and train them via the \textbf{Adversarial Loss}, allowing robust adaptation.

\noindent\textbf{Patch-level Rendering Loss.}
Secondly, the rendering loss is evaluated entirely within the highly non-linear RGB space, which suffers from ambiguity where identical colors appear at multiple spatial locations. Directly using pixel-level differences to supervise ray regression can lead to unstable optimization directions. To address this, we modify the rendering pipeline of NeRF. Instead of rendering only the patch center, we uniformly sample rays across the entire patch based on the predicted ray parameters. Then, the \textbf{Rendering Loss} is computed at the patch level, increasing its robustness compared to the pixel-level loss.

\noindent\textbf{Two-stage Training Pipeline.}
Additionally, RGB-guided pose optimization requires a good initial solution~\cite{posefromnerf}. 
We therefore formulate a {Two-stage Training Pipeline}. In the first stage, we solely use NeRF to render novel views for data augmentation. Once the localization model can predict rays reasonably well, NaCR initiates the second stage and back-propagates the rendering loss into the CRR model for fine-grained ray refinement.

Next, we elaborate our {Two-stage Training Pipeline} in detail.

\textit{1. NeRF Model Training and Rendering.} For each scene, we first train a \texttt{NeRFacto}~\cite{nerfstudio} model using the training images ($I\in\mathcal{I}_{real}$) and poses ($P\in\mathcal{P}_{real}$). 
We then create a set of novel camera poses $\mathcal{P}_{syn}$ by applying random translation and rotation noise  $\delta_t, \delta_r$.
Simultaneously, we perturb the image appearance embeddings~\cite{nerfstudio} to generate visual variations. 
Using these poses and appearance codes, we generate a synthetic image dataset ($\{\mathcal{I}_{syn}, \mathcal{P}_{syn}\}$).
To avoid introducing excessive noise from low-quality novel views, we filter the synthetic dataset using BRISQUE scores.

\textit{2. Training Stage 1 (\cref{fig:ts1}).} We train the CRR model on a 2:1 mixture of real and synthetic data using the GDT scheme. During each step, image patch features are passed to two heads: the Ray Regression Model that predicts rays and their confidence, and the Discriminator that classifies the feature domain. The model is optimized using only the Ray-Confidence loss on the regression output and the Adversarial loss driven by the Discriminator.

\textit{3. Training Stage 2 (\cref{fig:ts2}).} This stage introduces the Rendering Loss to provide direct image-space supervision. While retaining the data sampling strategy and loss functions from Stage 1, the predicted ray parameters are now also fed into the \texttt{NeRFacto} renderer. This enables the computation of the photometric loss between the rendered image patch and the ground-truth image patch. This loss is then back-propagated through \texttt{NeRFacto} and added to the existing Stage 1 objectives, jointly updating the Ray Regression Model.


Finally, benefiting from NeRF's rendering capabilities, we can feed the model's predicted poses into \texttt{NeRFacto} to render RGB and Depth. We then calculate the relative pose to the query image via 2D-3D matching to obtain a refined localization result, following~\cite{rap,grloc}, denoted as NaCR+NPR.

\subsection{Training Loss}

\subsubsection{\textbf{Ray-Confidence Loss}}
This loss optimizes the geometric accuracy of regressed rays. For each ray $r_i$, the training objective is:
\begin{equation}
L_{RC}=c^d_i||rd^{gt}_i - \hat{rd}_i||^2_2  + c^m_i||rm^{gt}_i - \hat{rm}_i||^2_2 - \alpha \log(c^m_ic^d_i)
\end{equation}
where $rd^{gt}_i,rm^{gt}_i$ are the ground-truth ray direction and moment parameters; $\hat{rd}_i,\hat{rm}_i$ are the prediction; $c_i^d,c_i^m$ are the predicted confidence; and $\alpha$ is a hyperparameter scaling the confidence regularization (typically set to 0.1).

\subsubsection{\textbf{Adversarial Loss}}
This loss encourages the feature extractor to generate domain-invariant features for both real and rendered images, training the Discriminator simultaneously. For any input image $I$ sampled from the sets of real ($\mathcal{I}_{real}$) or synthetic images ($\mathcal{I}_{syn}$), the loss function is:
\begin{equation}
L_{A} = |\mathcal{D}(\mathcal{E}(I)) - y|_1
\end{equation}
where $\mathcal{D}$ is the Discriminator, $\mathcal{E}$ represents the DINOv2 extractor specifically including the last two trainable attention layers, and $y$ is the domain label ($y=1$ if $I \in \mathcal{I}_{real}$, otherwise $y=0$).

\subsubsection{\textbf{Rendering Loss}}
The rendering loss is the photometric L2 distance between the color rendered by NeRF from a predicted ray and the ground-truth color of the input image patch.
For each ray with image center $(u,v)$ and patch size $s \times s$, the rendering loss is defined as:
\begin{equation}
    \mathcal{L}_{R} = \sum_{x=u- s/2 }^{u+ s/2 } \sum_{y=v- s/2 }^{v+ s/2 } \left\| \hat{\mathbf{I}}(x, y) - \mathbf{I}(x, y) \right\|^2_2
\end{equation}

where $\hat{\mathbf{I}}(x,y)$ is the image color rendered by \texttt{NeRFacto} at coordinates $(x,y)$, and $\mathbf{I}$ denotes the input image.

In summary, the total objective for Stage 1 optimization is $L = \beta L_{RC} + \gamma L_{A}$. For Stage 2 fine-tuning, the overall multi-task objective becomes $L = \beta L_{RC} + \gamma L_{A} + \mu L_{R}$. The $\beta, \gamma$ and $\mu$ are hyperparameters that balance the contribution of each loss term.

\section{Results}
\label{sec:res}

\begin{table*}[t]
\centering
\resizebox{0.85\linewidth}{!}{%
\begin{tabular}{c|c|c|c|c|c|c|c|c|c} 
\toprule
\multirow{2}{*}{Category} & \multirow{2}{*}{Method} & Chess & Fire & Heads & Office & Pumpkin & Kitchen & Stairs & Average \\
              &                 & ($cm$/$^{\circ}$) & ($cm$/$^{\circ}$) & ($cm$/$^{\circ}$) & ($cm$/$^{\circ}$) & ($cm$/$^{\circ}$) & ($cm$/$^{\circ}$) & ($cm$/$^{\circ}$) & ($cm$/$^{\circ}$) \\

\midrule
\multirow{3}{*}{SCR} 

& DSAC* \cite{dsac}              
& 0.5/{0.2}
& 0.8/0.3
& 0.5/0.3
& 1.2/0.3
& 1.2/0.3
& 0.7/0.2
& 2.7/0.8
& 1.1/0.3           \\

& ACE \cite{ace}          
& 0.5/0.2
& 0.8/0.3
& 0.6/0.3
& 1.0/{0.3}
& 1.0/0.2
& 0.8/0.2
& 2.9/0.8
& 1.1/0.3           \\

& GLACE \cite{GLACE}          
& 0.6/0.2
& 0.9/0.3
& 0.5/0.3
& 1.1/{0.3}
& 0.9/0.2
& 0.8/0.2
& 3.2/0.9
& 1.2/0.3           \\

\midrule

\multirow{2}{*}{RPR} & ExReNet \cite{ExReNet}              
& 5.0/1.6
& 7.0/2.5
& 3.0/2.7
& 6.0/1.8
& 7.0/2.0
& 7.0/2.1
& 19.0/4.9
& 8.0/2.5   \\

& Reloc3r \cite{reloc3r}              
& 3.0/0.9
& 3.0/0.8
& 1.0/1.0
& 4.0/0.9
& 6.0/1.1
& 4.0/1.3
& 7.0/1.3
& 4.0/1.0   \\

\midrule
\multirow{3}{*}{APR}

& $marepo$ \cite{maprepo}          
& 2.1/1.2
& 2.3/1.4
& 1.8/2.0
& 2.8/{1.3}
& 3.5/1.5
& 4.2/1.7
& 5.6/1.7
& 3.2/1.5           \\

& \ds RAP$_{GS}$ \cite{rap}          
& \ds {1.0}/0.8
& \ds 6.0/3.4
& \ds 4.0/5.5
& \ds 5.0/{1.9}
& \ds 4.0/1.7
& \ds 7.0/2.1
& \ds 9.0/2.1
& \ds 5.0/2.5           \\

& \ds GRLoc$_{GS}$ \cite{grloc}          
& \ds {1.0}/0.3
& \ds 2.9/1.0
& \ds 1.4/0.9
& \ds 5.1/{1.1}
& \ds 2.3/0.5
& \ds 3.1/0.7
& \ds 3.4/0.8
& \ds 2.8/0.8           \\


\midrule

\multirow{2}{*}{CRR}

&  DIMM~\cite{Zhang_2025_ICCV}
&  {{2.3}/0.9}
&  {2.6}/{1.2}
&  {{2.0}/1.3}
&  {{5.7}/1.0}
&  {2.7/0.8}
&  {3.2/0.8}
&  {{6.2}/1.1}
& 3.5/1.0 \\

& \ds NaCR$_{NeRF}$ (\textbf{Ours})
& \ds \boldblack{{0.9}/{0.2}}
& \ds \boldblack{{1.0}/{0.2}}
& \ds \boldblack{{1.3}/0.4}
& \ds \boldblack{{4.7}/0.6}
& \ds \boldblack{1.7/0.2}
& \ds \boldblack{1.9/0.3}
& \ds {\boldblack{3.8}/\boldblack{0.6}}
& \ds \boldblack{2.2/0.4}  \\

\midrule

\multirow{5}{*}{NPR}

& \ds CrossFire$_{NeRF}$ \cite{crossfire}          
& \ds {1.0}/0.4
& \ds 5.0/1.9
& \ds 3.0/2.3
& \ds 5.0/1.6
& \ds 3.0/0.8
& \ds 2.0/0.8
& \ds 12.0/1.9
& \ds 4.4/1.4           \\

& \ds NeRFMatch$_{NeRF}$ \cite{nerfmatch}          
& \ds {1.0}/0.3
& \ds 1.1/0.4
& \ds 1.3/0.9
& \ds 3.1/0.9
& \ds 2.2/0.6
& \ds 1.0/0.2
& \ds 9.3/1.7
& \ds 2.7/0.7           \\

& \ds RAP$_{GS}$ \cite{rap}          
& \ds {0.3}/0.1
& \ds 0.5/0.2
& \ds 0.4/0.3
& \ds {0.6}/\boldblack{0.1}
& \ds 0.8/0.2
& \ds 0.5/\boldblack{0.1}
& \ds 1.1/0.3
& \ds 0.6/0.2           \\

& \ds GRLoc$_{GS}$ \cite{grloc}          
& \ds {0.3}/\boldblack{0.1}
& \ds 0.3/\boldblack{0.1}
& \ds \boldblack{0.2}/\boldblack{0.1}
& \ds {0.6}/{0.2}
& \ds 0.8/0.2
& \ds 0.4/\boldblack{0.1}
& \ds  \boldblack{0.6}/ \boldblack{0.2}
& \ds 0.5/\boldblack{0.1}           \\

& \ds NaCR$_{NeRF}$ (\textbf{Ours})
& \ds {\boldblack{0.2}/\boldblack{0.1}}
& \ds \boldblack{0.2}/{0.6}
& \ds {{0.3}/\boldblack{0.1}}
& \ds {\boldblack{0.4}/0.3}
& \ds \boldblack{0.3/0.2}
& \ds \boldblack{0.2/0.1}
& \ds {{0.7}/ \boldblack{0.2}}
& \ds \boldblack{0.3}/0.2  \\

\bottomrule
\end{tabular}
}
\caption{{Localization accuracy on 7-Scenes~\cite{7scenes} using SfM poses as ground truth.} \normalfont We report median position errors in $cm$ and orientation errors in degrees ($^{\circ}$). The \boldblack{best} results within the CRR and NPR groups, together with \colorbox{colorSnd}{NVS-aided methods}, are highlighted. ``Method$_{NeRF/GS}$'' indicates that ``Method'' uses NeRF/GS-based novel view synthesis for data augmentation.}
\label{tab:7s}
\end{table*}

\subsection{Implementation Details}
For each scene, we construct a synthetic dataset by first training a \texttt{Nerfacto} model~\cite{nerfstudio} and subsequently rendering images from sampled novel-view poses. Detailed descriptions of the \texttt{Nerfacto} training process and the pose sampling strategy are provided in our Suppl.
The two-stage training of the NaCR takes place on an NVIDIA A800 GPU. To accommodate the pre-trained DA3 DINO backbone, we fix the image resolution to $504 \times 378$ and the patch size to $14$, predicting the ray parameters of the patch center. The parameters of the last two layers of the DA3 DINO backbone are unfrozen for updates. For our ray regression model, we adopt the DIMM decoder but replace its original \texttt{CLS} Token with the Camera Token from DA3 DINOv2 backbone, which also sufficiently summarizes the global image content.


For the first training stage, we employ the AdamW optimizer with a learning rate set to $10^{-4}$ and weight decay at $0.01$, utilizing GDT for 40K steps. In the second stage, we incorporate the patch-level rendering loss and lower the learning rate empirically to $10^{-6}$ for another 20K steps. A batch size of 5120 is consistently used during training. Empirically, the loss parameters are set to: $\beta=1$, $\gamma=0.2$, and $\mu=0.1$.

\subsection{7-Scenes Results}

\begin{table*}[t]
\centering
\resizebox{\linewidth}{!}{
\footnotesize
\begin{tabular}{l|cc|cccc|c|cc}
\toprule
\multirow{4}{*}{Scene}& \multicolumn{2}{c}{SCR}  &  \multicolumn{4}{c}{APR}   &  \multicolumn{3}{c}{CRR} \\ \cmidrule(l){2-3} \cmidrule(l){4-7} \cmidrule(l){8-10}
& DSAC* & ACE & PN & MST & $marepo$ & $marepo_{S}$ &  DIMM & \ours NaCR &  \ours NaCR+NPR  \\
& \cite{dsac} & \cite{ace} & \cite{pn0,pn1,pn2} & \cite{mst} & \cite{maprepo} & \cite{maprepo} &  \cite{Zhang_2025_ICCV} &   \ours Ours &   \ours Ours \\
\midrule
Throughput (fps)    &   17.9       &    17.9 &  {166.7}      &      28.4         &     55.6     & 55.6  &  20.1    &  \ours 33.4 &  \ours 16.5    \\
\midrule
Bears        & 82.6\%/91.6\%  & 80.7\%/92.6\%  & 12.9\%/35.7\% & 0.5\%/12.8\%   & {80.7\%}/99.3\%             & {80.7\%}/{99.5\%}  &  {82.4}\%/97.1\% &  \ours 95.5\%/99.3\% &   \ours \textbf{96.4}\%/\textbf{99.7}\% 
\\
Cubes        & 83.8\%/98.1\%  & \textbf{97.0}\%/\textbf{98.1}\%  & 0.0\%/0.4\%    & 0.00\%/9.9\%  & {72.4\%}/{96.9\%} & 71.8\%/{96.9\%} &  71.5\%/{89.8}\% &  \ours 72.5\%/85.7\%  &  \ours {73.8}\%/87.9\%
\\
Inscription  & {54.1}\%/69.7\%  & 49.0\%/69.6\%  & 1.1\%/6.3\%   & 1.3\%/9.7\%    & {37.8\%}/{74.2\%} & 37.1\%/74.1\% &  21.3\%/{72.6}\% &   \ours \textbf{57.8}\%/75.1\% &   \ours 56.1\%/\textbf{77.3}\% 
\\
Lawn         & 34.7\%/38.0\%    & \textbf{35.8}\%/38.5\%  & 0.0\%/0.2\%    & 0.0\%/0.0\%    & 32.6\%/{41.6\%}          & {34.2\%}/41.1\% &  22.4\%/{46.3}\% &   \ours 34.2\%/47.5\% &   \ours {35.1}\%/\textbf{48.8}\% 
\\
Map          & 56.7\%/87.1\%  & 56.5\%/84.7\%  & 14.9\%/49.1\% & 5.6\%/25.7\%   & 53.9\%/87.7\%                         & {55.1\%}/{87.9\%} &  {57.3}\%/{89.2}\% &   \ours 68.0\%/95.2\% &   \ours \textbf{68.7}\%/\textbf{96.3}\% 
\\
Square Bench & 69.5\%/97.9\%  & 66.7\%/97.8\%  & 0.0\%/3.0\%    & 0.0\%/0.0\%   & 68.6\%/\textbf{100\%}              & {70.7\%}/\textbf{100\%}  &  54.3\%/98.9\% &   \ours 70.9\%/97.1\% &   \ours \textbf{72.3}\%/98.4\%
\\
Statue       & 0.0\%/0.0\%     & 0.0\%/0.0\%     & 0.0\%/0.0\%     & 0.0\%/0.0\%    & 0.0\%/0.0\%                       & 0.0\%/0.0\%   &  0.0\%/0.0\%   &   \ours \textbf{1.3}\%/\textbf{4.1}\%  &   \ours 0.5\%/2.2\% 
\\
Tendrils     & 25.1\%/26.5\%  &34.9\%/36.8\%   & 0.0\%/0.0\%     & 0.9\%/23.6\% & 27.9\%/33.4\%                         & {29.3\%}/{34.8\%} &  24.1\%/{43.6}\% &   \ours 46.8\%/50.4\% &   \ours \textbf{47.3}\%/\textbf{52.6}\% 
\\
The Rock     & \textbf{100}\%/\textbf{100}\%      & \textbf{100}\% /\textbf{100}\%      & 24.2\%/77.5\% & 10.7\%/52.6\% & 98.1\%/\textbf{100}\%                &{99.8}\%/\textbf{100}\%   &  95.4\%/\textbf{100}\%    &   \ours 97.9\%/\textbf{100}\%  &   \ours 98.6\%/\textbf{100}\% 
\\
Winter Sign  & 0.2\%/5.7\%    & \textbf{1.0}\%/7.6\%    & 0.0\%/0.0\%     & 0.0\%/0.0\%  & 0.0\%/{0.7\%}               & 0.0\%/0.3\%   &  0.0\%/{5.4}\% &   \ours 0.2\% \textbf{8.3}\% &   \ours 0.0\% {7.6}\% 
\\
\midrule
Average      & 50.7\%/61.5\%  & {52.2}\%/62.6\%  & 5.3\%/17.2\%  & 1.9\%/13.4\%   & 47.2\%/63.4\%                         & {47.9\%}/{63.5\%} &  {42.9\%}/{64.3\%} &   \ours {54.5\%}/{66.3\%} &   \ours \textbf{54.9}\%/\textbf{67.1}\% 
\\
\bottomrule
\end{tabular}
}
\caption{\textbf{Pose accuracy comparison on the outdoor Wayspots~\cite{mapfree} dataset.}
\normalfont Results are reported as the percentage of frames below $10cm/5^\circ $ and $0.5m/5^\circ$ error.
}
\label{tab:way}
\end{table*}

\noindent\textbf{Dataset.}
We first evaluate the localization accuracy on the indoor 7-Scenes~\cite{7scenes} dataset. This dataset provides seven indoor scenes, each containing several video sequences captured from different trajectories. The resolution of images is low and features serious motion blur. We follow ACE~\cite{ace} to train and test with accurate SfM poses.

\noindent\textbf{Baselines.}
On this dataset, we selected methods from several different paradigms for comparisons: Scene Coordinate Regression (SCR) paradigm including DSAC*~\cite{dsac}, ACE~\cite{ace}, and GLACE~\cite{GLACE}; methods based on Absolute Pose Regression (APR)~\cite{maprepo} and Relative Pose Regression (RPR)~\cite{ExReNet,reloc3r}; RAP~\cite{rap} and GRLoc~\cite{grloc} which apply 3DGS for data augmentation; and finally, DIMM~\cite{Zhang_2025_ICCV} under the Camera Ray Regression (CRR) paradigm. Besides, the NVS models can assist in post-inference optimization~\cite{crossfire,nerfmatch}, which is also feasible in our method. Following RAP, we denote these methods as \colorbox{colorSnd}{NVS-based Pose Refinement (NPR)} and compare their performances. We report the median translation and rotation errors across scenes.

\noindent\textbf{Results.}
As shown in \cref{tab:7s}, NaCR substantially improves accuracy over DIMM, reducing the average translation error from $3.5$~cm to $2.2$~cm. This improvement comes from both incorporating NeRF and improving the baseline. Ablation studies of these two modules are presented in \cref{tab:mapping_time} and \cref{tab:ablation_dimm} of the supplementary material. Compared with methods adopting NVS models for data augmentation, \ie, RAP and GRLoc, NaCR also yields superior accuracy (\eg, GRLoc's translation error is up to $2.8$~cm, while NaCR reduces it to $2.2$~cm). These results highlight the primary advantage of NaCR: applying an NVS model for both data augmentation and in-training supervision yields better performance than augmentation alone.
While NaCR trails the top SCR-based methods on this dataset, this is expected because the bounded indoor environment minimizes SCR's weakness, namely, instability from unbounded background points. However, leveraging NaCR's initial pose estimates, NaCR+NPR achieves a $0.3$~cm translation error, surpassing both SCR and other post-optimization methods. The qualitative results are presented in Figure~\ref{fig:q7s}.

\subsection{Wayspots Results}

\noindent\textbf{Dataset.}
We also conduct experiments on the outdoor Wayspots dataset~\cite{mapfree}. Wayspots involves 10 outdoor scenes, each enclosing two video sequences dedicated separately to mapping and localization. Due to low-texture and repetitive areas within views, this dataset provides remarkable challenges for visual localization.

\noindent\textbf{Baselines.}
In this experiment, we compare against SCR methods specifically DSAC*~\cite{dsac} and ACE~\cite{ace}; APR methods including PN~\cite{pn0,pn1,pn2}, MST~\cite{mst}, and marepo~\cite{maprepo}; as well as the CRR baseline DIMM~\cite{Zhang_2025_ICCV}. Performance is measured using localization recall at $10{cm}/5^\circ$ and $0.5{m}/5^\circ$ error margins.

\begin{table}[!t]
    \centering
    \caption{Detailed efficiency report. \normalfont Mapping time is measured per scene, and inference time is measured per query image.}
    \label{tab:efficiency_report}
    \resizebox{0.7\linewidth}{!}{
    \begin{tabular}{lccc}
        \toprule
        Component & Model Params (M) & Time & Role \\
        \midrule
        ACE Mapping & 2.10 & 5 min & SCR mapping \\
        GLACE Mapping & 9.45 & 33 min & SCR mapping \\
        DIMM Mapping & 18.31 & 1.32 h & baseline mapping \\
        \textbf{NaCR Mapping Cost} & 39.53 & 3.63 h & mapping \\
        \quad$\hookrightarrow$~Scene NeRF training & 16.79 & 0.42 h & mapping \\
        \quad$\hookrightarrow$~Stage-1 training & 39.53 & 2.08 h & mapping \\
        \quad$\hookrightarrow$~Stage-2 training & 39.53 & 1.13 h & mapping \\
        \midrule
        DIMM Inference & 18.31 & 0.02 s & baseline VL \\
        \textbf{NaCR Inference Cost} & 39.53 & 0.03 s & VL \\
        \bottomrule
    \end{tabular}
    }
\end{table}

\noindent\textbf{Results.}
As demonstrated in \cref{tab:way}, NaCR yields \textit{state-of-the-art} VL results across scenes. The integration of NeRF into CRR results in an overall precision boost. NaCR reaches $54.5\%$ for the $10\text{cm}/5^\circ$ recall metric, comparing favorably with the $42.9\%$ of baseline DIMM. Unlike SCR methods, NaCR exhibits high robustness to the unbounded backgrounds of the Wayspots dataset. This allows NaCR ($54.5\%$) to outperform ACE ($52.2\%$). Our NeRF-based post-optimization step (NaCR+NPR) generally provides further gains in most scenes. However, we also observe degraded optimization in a few scenes, such as ``State'' and ``Winter Sign''. This is likely due to the prevalence of repetitive textures in these environments, which reduces matching accuracy and introduces errors into the optimization process. Qualitative results are provided in the supplementary material~\ref{sec:suppl_qws}.

\subsection{Efficiency Report}

We report the cost of each NaCR component in \cref{tab:efficiency_report}. The NeRF is used as a one-time offline mapping module: it is trained once per scene and then provides both synthetic views and differentiable rendering supervision during NaCR training. This increases the per-scene mapping time from 1.32 h for DIMM to 3.63 h for NaCR, including 0.42 h for scene NeRF training, 2.08 h for stage-1 training, and 1.13 h for stage-2 training. At localization time, NaCR keeps the feed-forward CRR inference path and does not require NeRF optimization, resulting in a comparable per-query runtime of 0.03 s. As a complementary analysis, Suppl.~D provides a Mapping Time Profile that studies the accuracy--cost trade-off under different data amounts of rendering augmentation.

\subsection{Ablation Study on Training Pipeline}

We ablate the two-stage strategy on 7-Scenes by comparing single-stage training with and without the rendering loss. As summarized in \cref{tab:component_ablation}, applying the rendering loss from the start destabilizes optimization (see \cref{fig:ts} in the Suppl.) and degrades accuracy, while removing it is also sub-optimal. The two-stage pipeline is therefore critical: it first stabilizes CRR with GDT, then uses the rendering loss for local ray refinement.

\subsection{Complete Component Ablation}
\label{sec:component_ablation}

We further isolate the contributions of the architectural upgrades, NeRF-based data augmentation, and patch-level rendering loss on the full 7-Scenes and Wayspots benchmarks. Table~\ref{tab:component_ablation} reports results averaged over three random seeds. The matched pair NaCR ($\mu=0$) and NaCR differs only in the rendering-loss weight and therefore directly measures the effect of $\mathcal{L}_R$ under the same two-stage schedule. The rendering loss improves the full model from $2.4/0.5$ to $2.2/0.4$ on 7-Scenes and from $52.7/65.5$ to $54.1/66.1$ on Wayspots, while the single-stage variant with rendering loss is substantially less stable. These results support the complementary roles of the three components.

\begin{table}[t]
    \centering
    \small
    \setlength{\tabcolsep}{3.2pt}
    \renewcommand{\arraystretch}{1.08}
    \caption{Complete component ablation. Except for the two single-stage variants, all configurations use a 60K-step two-stage schedule. Results are averaged over seeds 42, 23, and 24 and across all scenes in 7-Scenes and Wayspots. ``Arch.'' denotes the architectural upgrades. Each entry reports translation/rotation error on 7-Scenes and recall at $10\text{cm}/5^\circ$ and $0.5\text{m}/5^\circ$ on Wayspots.}
    \label{tab:component_ablation}
    \resizebox{\linewidth}{!}{%
    \begin{tabular}{lcccccccc}
        \toprule
        Method & Arch. & NVS aug. & $\mathcal{L}_{R}$ & TST & 7-Scenes $\downarrow$ & STD & Wayspots $\uparrow$ & STD \\
        \midrule
        DIMM & $\times$ & $\times$ & $\times$ & $\checkmark$ & 3.6/1.0 & 0.034 & 42.3/64.0 & 0.215 \\
        \quad$\hookrightarrow$ DA3 only & DA3 & $\times$ & $\times$ & $\checkmark$ & 3.3/0.9 & 0.037 & 45.9/63.8 & 0.052 \\
        DIMM$^{*}$ & Full & $\times$ & $\times$ & $\checkmark$ & 3.2/0.9 & 0.019 & 45.8/63.7 & 0.555 \\
        NaCR ($\mu=0$) & Full & $\checkmark$ & $\times$ & $\checkmark$ & 2.4/0.5 & 0.022 & 52.7/65.5 & 0.040 \\
        DIMM$^{*}$ + $\mathcal{L}_{R}$ & Full & $\times$ & $\checkmark$ & $\checkmark$ & 2.9/0.7 & 0.023 & 51.0/64.3 & 0.117 \\
        NaCR-SST & Full & $\checkmark$ & $\checkmark$ & $\times$ & 6.7/3.3 & 0.027 & 39.1/60.0 & 0.099 \\
        NaCR-SST ($\mu=0$) & Full & $\checkmark$ & $\times$ & $\times$ & 2.8/0.7 & 0.048 & 50.2/64.3 & 0.119 \\
        \textbf{NaCR} & Full & $\checkmark$ & $\checkmark$ & $\checkmark$ & \textbf{2.2/0.4} & 0.033 & \textbf{54.1/66.1} & 0.156 \\
        \bottomrule
    \end{tabular}%
    }
\end{table}

\section{Conclusion}
\label{sec:conc}

This work proposes a novel Camera Ray Regression method, NaCR, for Visual Localization.
Building upon a previous baseline, NaCR first introduces three simple yet effective improvements inspired by advanced camera ray-based reconstruction methods.
The core technical insight of NaCR is the deep integration of NeRF and CRR, driven by the observation that camera rays hold a central role in both techniques.
Specifically, NaCR exploits NeRF's novel view synthesis capability to generate abundant pose-image pairs for ray-level synthetic data augmentation.
Furthermore, taking advantage of the differentiable rendering of NeRF, NaCR constructs a closed-loop gradient flow between the image and ray space, providing photometric supervision for ray regression.
To overcome the training instability caused by the highly non-linear nature of the image space, we design a stable two-stage training scheme, leveraging rendering errors to fine-tune the rays locally while ensuring optimization stability.
Extensive experiments on both indoor and outdoor datasets demonstrate that NaCR achieves competitive performance. Our ablation studies thoroughly validate the contribution of each module. 

A limitation of the current work is its reliance on a pre-trained NeRF for each scene, which increases the offline mapping cost and may limit scalability to large or changing environments. Nevertheless, this reliance may also be viewed not merely as a drawback, but as a potentially useful design choice. This ray-level interface also provides a natural path for future extension, as it can readily incorporate advances in NeRF designed for large-scale environments. In future work, we plan to build on this design to improve scalability and reduce the need for scene-specific NeRF preparation. More details can be found in our Suppl.

\bibliographystyle{splncs04}
\bibliography{main}

\appendix
\clearpage
\section*{Supplementary Material}

\section{Qualitative Results on Wayspots}
\label{sec:suppl_qws}

\begin{figure*}[!h]
    \centering
    \includegraphics[width=0.8\linewidth]{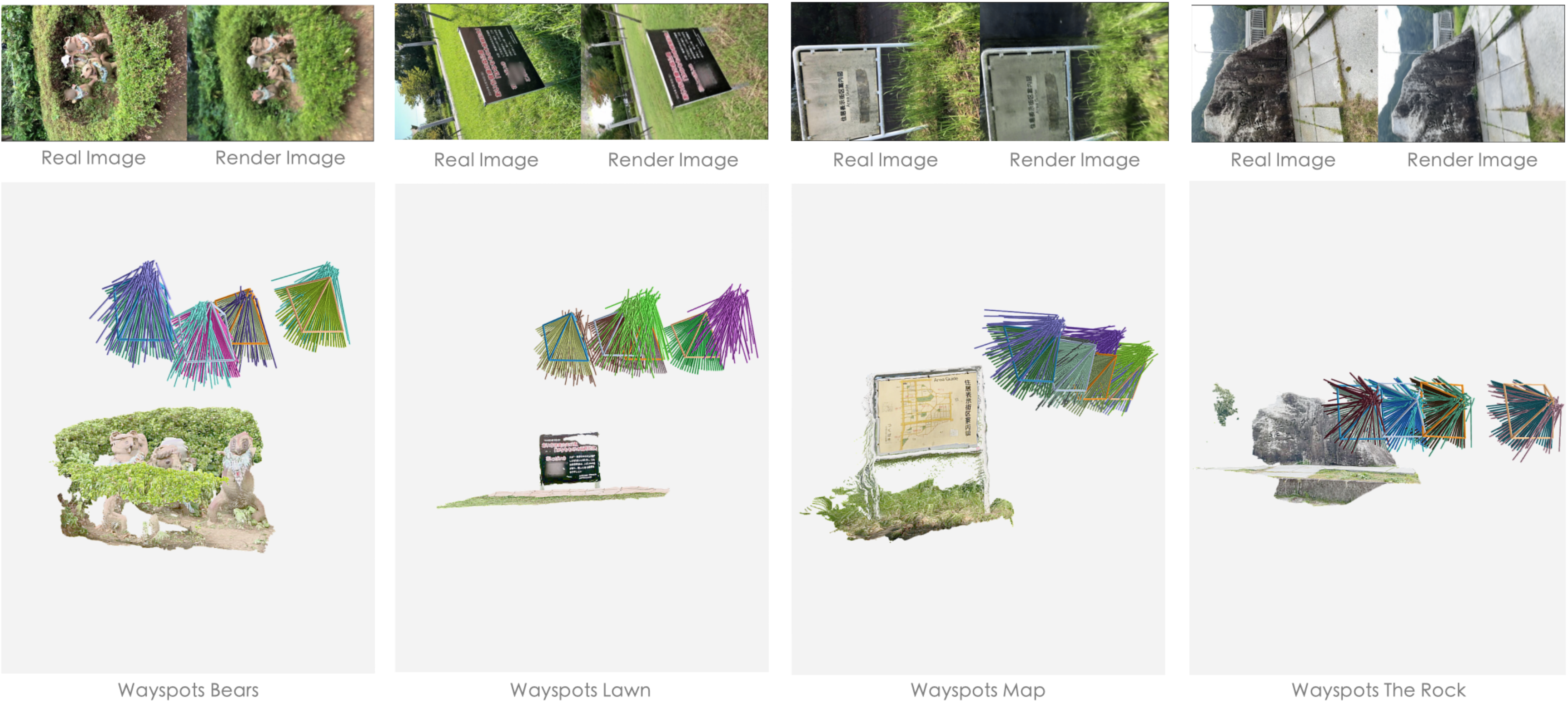}
    \caption{Qualitative results in Wayspots.\normalfont The top row compares the ground truth image with a view rendered from our estimated camera pose. The bottom row visualizes the predicted camera rays within the 3D scene.}
    \label{fig:qws}
    \vspace{-0.2em}
\end{figure*}

We provide qualitative results on the outdoor Wayspots dataset in \cref{fig:qws}. The top row compares each query image with the view rendered from our estimated camera pose. The close visual agreement indicates that the predicted pose is geometrically consistent with the scene representation learned by \texttt{Nerfacto}, even under outdoor appearance changes and repetitive structures. The bottom row further visualizes the predicted camera rays in the reconstructed scene. These ray visualizations show that NaCR produces spatially coherent ray predictions rather than only fitting the final pose after aggregation, supporting the effectiveness of coupling CRR with NeRF at the ray level.

\section{Implementation Details of Synthesis Data Augmentation}

For each scene, we train a \texttt{Nerfacto} model~\cite{nerfstudio} using the Adam optimizer for 20K steps on an NVIDIA RTX 4090 GPU, with a learning rate set to 0.01. Then, we use the \texttt{Nerfacto} model to generate a novel view dataset: $\{\mathcal{I}_{syn}, \mathcal{P}_{syn}\}$. Specifically, we compute the variation range in translation and rotation for the poses in the training set ($\mathcal{P}_{real}$), and compute the maximum variations for translation and rotation: $\Delta_{t}, \Delta_{r}$, respectively. Then, the novel view poses ($P_{syn} \in \mathcal{P}_{syn}$) are generated based on random perturbations $\delta_t \in (0, \Delta_t)$ and $\delta_r \in (0, \Delta_r)$ on the real poses. We render images at these poses and filter them using a BRISQUE score threshold of 50, resampling any low-quality views.


\section{Ablation Study on Adversarial Loss}

\begin{table}[ht]
    \centering
    \caption{Ablation Study on the adversarial loss. \normalfont The median rotation and translation errors are reported.}
    \label{tab:ablation_adv}
    \resizebox{0.4\linewidth}{!}{
    \begin{tabular}{l c c}
        \toprule
        Scene & NaCR & \makecell{NaCR \\ \small w/o Adversarial Loss} \\
        \midrule
        chess  & \textbf{0.9/0.2} & 1.0/0.3 \\
        fire   & \textbf{1.0/0.2} & 1.2/0.5 \\
        heads  & \textbf{1.3/0.4} & 1.7/0.6 \\
        office & \textbf{4.7/0.6} & 4.9/0.9 \\
        \bottomrule
    \end{tabular}
    }
\end{table}

To evaluate the effectiveness of the adversarial loss, which is designed to bridge the render-to-real domain gap, we conducted an ablation study on the 7-Scenes dataset. The results, presented in Table~\ref{tab:ablation_adv}, show that removing the adversarial component leads to a consistent increase in localization error. This confirms that adversarial training acts as a crucial regularizer, making our feature extractor robust to the domain shift between synthetic and real data.

\begin{figure}[!ht]
    \centering
    \includegraphics[width=\linewidth]{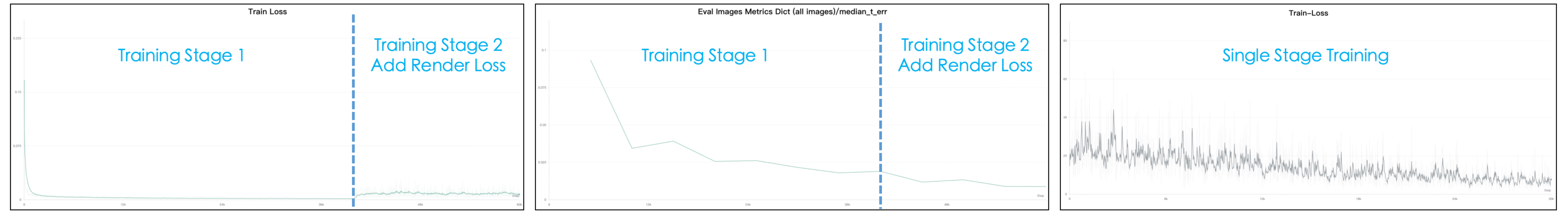} 
    \caption{Training loss curves on the 7-Scenes Chess dataset, comparing our two-stage training strategy against a single-stage baseline. \normalfont Our two-stage training demonstrates that while the rendering loss introduces slight instability to the training error~(left), it yields superior final translation accuracy~(middle). Conversely, the single-stage baseline~(right), which applies the rendering loss from the start, struggles to converge with optimization instability, due to the unreliable supervision signal from the renderer when ray predictions are still coarse.}
    \label{fig:ts}
\end{figure}

\section{Mapping Time Profile}
\label{sec:mt}

\begin{table}[ht]
    \centering
    \caption{Mapping time versus performance. \normalfont The $10\text{cm}/5^\circ$ recall $\uparrow$ and training steps are reported on Wayspots-Bears, varying the proportion of augmented synthesized images.}
    \label{tab:mapping_time}
    \resizebox{0.6\linewidth}{!}{
    \begin{tabular}{cccc}
        \toprule
        \makecell{Proportion of \\ \small Synthesized Images} & Recall ($\%$) & Time (h) & Training Steps \\
        \midrule
        DIMM (Baseline) & 82.4 & \textbf{0.83} & 20K \\
        0\% (Improved DIMM)  & 84.1 & 1.15 & 20K \\
        20\%            & 84.6 & 1.87 & 30K \\
        50\%            & 87.1 & 2.25 & 40K \\
        70\%            & 92.3 & 2.67 & 50K \\
        100\%           & \textbf{95.5} & 3.22 & 60K \\
        \bottomrule
    \end{tabular}
    }
\end{table}

This section investigates the trade-off between training duration and model performance relative to the volume of synthetic data utilized. As the training steps are increased, total training time scales linearly with the amount of rendered data added for augmentation. To quantify this relationship, we evaluated VL recall on the Wayspots-Bears dataset across varying ratios of synthetic data, with results summarized in Table \ref{tab:mapping_time}. Our findings indicate that increasing the volume of augmented data yields progressively higher accuracy, albeit at the expense of longer training periods. This clear trade-off enables the selection of an optimal augmentation level that balances predictive accuracy against computational overhead. Furthermore, in the absence of synthetic data, our improved DIMM outperforms the original DIMM, demonstrating the inherent effectiveness of the proposed architectural enhancements over the baseline model.

\section{Augmentation from Different NVS Models}
\label{sec:suppl_nvs_aug}

\begin{table}[!t]
    \centering
    \caption{Ablation study on NeRF augmentation. \normalfont The median translation/rotation errors $\downarrow$, along with rendering PSNR $\uparrow$, are reported on four 7-Scenes scenes.}
    \label{tab:ablation_nerf}
    \resizebox{0.72\linewidth}{!}{
    \begin{tabular}{l ccccc}
        \toprule
        Scene & NaCR & NaCR & NaCR & RAP~\cite{rap} & GRLoc~\cite{grloc} \\
        \cmidrule(lr){2-4} \cmidrule(lr){5-6}
        NVS model & Nerfacto~\cite{nerfstudio} & Deblur-NeRF~\cite{deblurnerf} & \makecell{3DGS~\cite{ye2025gsplat} \\ \small (w/o rendering loss)} & 3DGS & 3DGS \\
        \midrule
        chess  & \textbf{0.9}/\textbf{0.2}/23.67 & 1.0/\textbf{0.2}/22.91 & 1.1/0.5/\textbf{23.95} & 1.0/0.8/- & 1.0/0.3/- \\
        fire   & \textbf{1.0}/\textbf{0.2}/22.75 & 1.1/0.5/21.33 & 1.4/0.3/\textbf{23.86} & 6.0/3.4/- & 2.9/1.0/- \\
        heads  & \textbf{1.3}/\textbf{0.4}/18.89 & 1.5/1.0/18.09 & 1.7/1.3/\textbf{19.07} & 4.0/5.5/- & 1.4/0.9/- \\
        office & {4.7}/\textbf{0.6}/{21.98} & \textbf{3.2}/0.8/\textbf{22.85} & 5.3/1.3/22.69 & 5.1/1.1/- & 5.0/1.9/- \\
        \bottomrule
    \end{tabular}
    }
\vspace{-0.3em}
\end{table}

As NaCR embeds an NVS model into CRR, we ablate NVS choices on the first four motion-blurred 7-Scenes scenes. We compare our default \texttt{Nerfacto}~\cite{nerfstudio}, Deblur-NeRF~\cite{deblurnerf}, and \texttt{Splatfacto}/3DGS~\cite{ye2025gsplat}; the last provides high-quality augmentation but lacks the explicit ray geometry required by our rendering loss. \cref{tab:ablation_nerf} also reports PSNR and APR baselines with 3DGS augmentation. \texttt{Nerfacto} gives the best average localization, while Deblur-NeRF helps only the heavily blurred ``office'' scene. Although 3DGS achieves the highest PSNR, the missing rendering loss limits localization accuracy. Across all NVS choices, NaCR outperforms APR methods with NVS augmentation, supporting both the CRR paradigm and tight NVS-CRR integration.

\section{NVS Reference-View Sensitivity}
\label{sec:reference_view_sensitivity}

To quantify the effect of NVS training coverage, we vary the number of reference views used to train the scene representation from 25\% to 100\% of the original training images. Nerfacto and 3DGS use the same reference-view subsets at every setting. Table~\ref{tab:reference_view_sensitivity} reports both test-view rendering quality and downstream localization accuracy. NaCR improves steadily as reference-view coverage increases, even though Nerfacto has lower image-level PSNR than 3DGS. This trend is consistent with patch-level ray supervision: local rendering signals can still provide useful gradients for ray refinement, and better NVS coverage yields more reliable supervision.

\begin{table}[t]
    \centering
    \scriptsize
    \setlength{\tabcolsep}{3.2pt}
    \renewcommand{\arraystretch}{1.08}
    \caption{NVS reference-view sensitivity on 7-Scenes. Nerfacto and 3DGS use matched reference-view subsets. We report average median translation/rotation error (cm/$^\circ$) and test-view PSNR over all scenes.}
    \label{tab:reference_view_sensitivity}
    \resizebox{\linewidth}{!}{%
    \begin{tabular}{lccccc}
        \toprule
        NVS reference views (\%) & NaCR $\downarrow$ & Nerfacto PSNR $\uparrow$ & RAP $\downarrow$ & 3DGS PSNR $\uparrow$ \\
        \midrule
        0   & 2.9/0.7 & N/A  & 7.1/2.9 & N/A  \\
        25  & 3.2/1.1 & 13.7 & 8.3/3.6 & 15.3 \\
        50  & 2.6/0.6 & 17.4 & 7.5/3.2 & 19.4 \\
        75  & 2.3/0.4 & 19.8 & 6.8/2.8 & 22.3 \\
        100 & \textbf{2.2/0.4} & 21.0 & 5.0/2.5 & 23.4 \\
        \bottomrule
    \end{tabular}%
    }
\end{table}

\section{Ablation Study on Improved DIMM}
\label{sec:ab_dimm}

\begin{table}[t]
    \centering
    \caption{Ablation Study on Improved DIMM components. \normalfont The recall $\uparrow$ within $10\text{cm}/5^\circ$ is reported on the Wayspots dataset.}
    \label{tab:ablation_dimm}
    \resizebox{0.6\linewidth}{!}{
    \begin{tabular}{lcccc}
        \toprule
        Model & Bears & Cubes & Lawn & The Rock \\
        \midrule
        I: DIMM                      & 82.4 & 71.5 & 22.4 & 95.4 \\
        II: DIMM + \texttt{Nerfacto}          & 91.7 & 72.1 & 29.6 & 96.0 \\
        III: II + DA3 DINOv2         & 92.5 & 72.3 & 30.2 & 96.5 \\
        \makecell[l]{IV: III + Ref.~Token~ Cond. \small (random)} & 92.1 & 72.1 & 29.8 & 96.9 \\
        \makecell[l]{V: III + Ref.~Token~ Cond. \small (centered)} & 94.6 & 72.3 & 32.5 & \textbf{98.1} \\
        VI: V + Ray-Confidence Head  & \textbf{95.5} & \textbf{72.5} & \textbf{34.2} & {97.9} \\
        \bottomrule
    \end{tabular}
    }
\end{table}

In this section, an ablation study is conducted on four scenes from the Wayspots dataset to validate our three proposed enhancements to the DIMM baseline, using recall at $10\text{cm}/5^{\circ}$ as the metric.
The results in \cref{tab:ablation_dimm} confirm that every component provides a distinct benefit.
Specifically, adopting the pre-trained DINOv2 backbone from DA3 consistently improves performance, such as $91.7$ vs. $92.5$ in the ``Bears'' scene.
For the reference token condition, we found that our pose-clustering method is superior to random sampling, which can degrade accuracy.
This suggests that a well-distributed set of reference images can produce more robust query features.
Finally, the inclusion of the ray confidence head also boosts performance in most cases, except for low-complexity environments such as ``The Rock''.
Furthermore, a cross-table comparison reveals that the NeRF integration (\cref{tab:ablation_dimm}: DIMM vs. DIMM + \texttt{Nerfacto}) provides a greater performance contribution than our architectural enhancements to the baseline (\cref{tab:mapping_time}: DIMM vs. Improved DIMM).
However, both of them are significant and necessary to achieve the final performance of NaCR.

\section{Fine-tuning DPT of DA3}

\begin{figure*}[!t] 
    \centering
    \includegraphics[width=\linewidth]{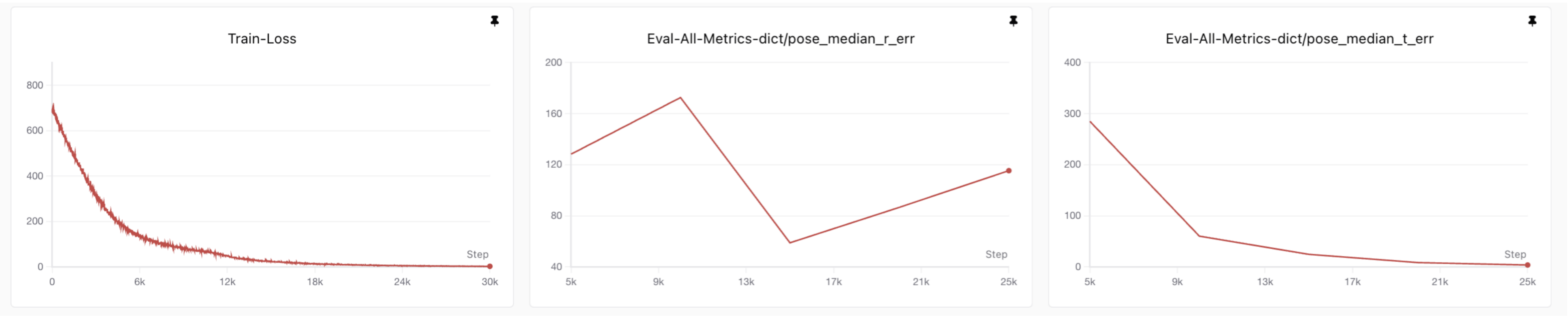} 
    \caption{The performance of image-level fine-tuning of DA3 DPT~\cite{dpt} decoder in 7-Scenes Chess. \normalfont This is because the small VL training set generates highly correlated ray gradients within each image, which corrupts the powerful geometric priors learned during DA3's large-scale pre-training.}
    \label{fig:sdpt}
\end{figure*}

We also evaluate an alternative training approach: directly fine-tuning the original DA3 Dense Predictive Transformer~\cite{dpt} (DPT) decoder at the image level, instead of using the patch-based DIMM decoder with GDT. However, this strategy yields poor localization performance, as shown in \cref{fig:sdpt}. We attribute this failure to two primary factors: \textit{1) Catastrophic Forgetting:} The training data for a single VL scene is vastly smaller and less diverse than DA3's pre-training dataset. Fine-tuning on this narrow distribution can corrupt the powerful, generalizable geometric priors of DPT acquired during large-scale pre-training. \textit{2) Training Inefficiency:} Image-level training is inefficient because the gradients from the rays within a single image are highly correlated. This lack of sample diversity per step significantly slows convergence on the single scene and limits the VL accuracy~\cite{ace}.
Given these limitations, the original image-level decoder is ill-suited for the scene-specific VL task. We therefore adopt the more robust and efficient combination of a patch-level DIMM decoder~\cite{Zhang_2025_ICCV} and the GDT scheme~\cite{ace}.

\section{Limitation and Future Work}

\begin{figure*}[!t] 
    \centering
    \includegraphics[width=0.8\linewidth]{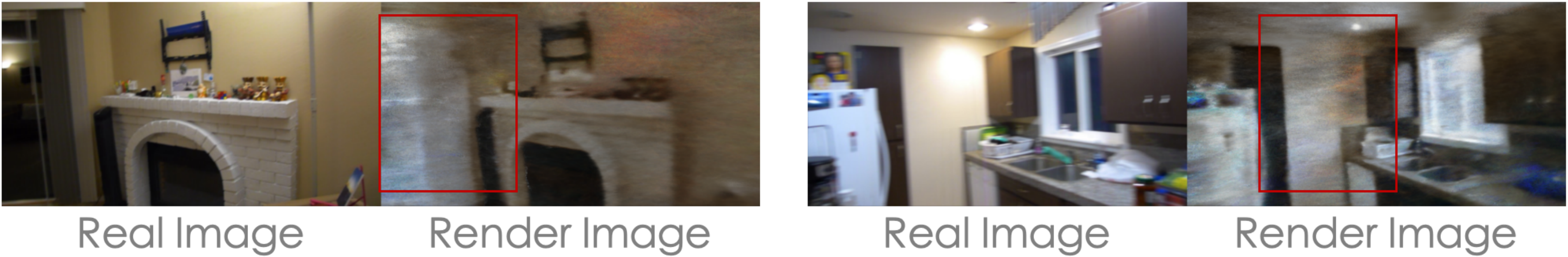} 
    \caption{Poor Rendering Image Quality of \texttt{Nerfacto} in 6-Scenes~\cite{sld}.\normalfont The severe illumination changes in the 6-Scenes dataset degrade the rendering quality of the NVS model, causing it to produce significant visual artifacts. This, in turn, negatively impacts NaCR's localization performance by introducing noise into both of its NeRF-dependent components: (1) the synthetic images used for data augmentation are corrupted, and (2) the supervisory signal from the end-to-end rendering loss becomes unreliable.}
    \label{fig:6s}
\end{figure*}

\subsection{Limitation from NeRF Reliance}
A key limitation of NaCR is its dependence on a high-quality, pre-trained NeRF for each scene. 
In challenging localization scenarios suffering from sparse image coverage or severe illumination changes, reconstructing a high-quality NeRF remains difficult. Therefore, the inherent noise within the NeRF model can degrade the VL performance of NaCR. For instance, we attempt to apply NaCR to the 6-Scenes dataset~\cite{sld}, which is characterized by large spatial ranges and lighting variations. 
These conditions lead to a low-quality NeRF reconstruction containing significant visual artifacts (as shown in \cref{fig:6s}). When integrated into the NaCR pipeline, these rendering errors introduce substantial noise into both the data augmentation and the supervisory signal, ultimately causing NaCR's VL performance to fall below that of the DIMM baseline.

\subsection{Future work of Jointly NeRF-CRR Optimization}
A promising direction for future work is the end-to-end, joint optimization of the NeRF and CRR models. 
While our differentiable pipeline theoretically supports this, it presents a significant challenge: NeRF training relies on accurate camera rays, while our CRR model requires a high-fidelity NeRF for supervision. Successfully resolving this co-dependency could lead to a framework capable of simultaneous localization and mapping. This function is similar to recent Large View Synthesis Models (LVSM)~\cite{rayzer,erayzer}, but is explicitly grounded in geometry of the camera ray, potentially offering greater robustness and interpretability.

\end{document}